\documentclass{article}

\usepackage[final, nonatbib]{neurips_2021}

\usepackage[utf8]{inputenc} 
\usepackage[T1]{fontenc}    
\usepackage{hyperref}       
\usepackage{url}            
\usepackage{booktabs}       
\usepackage{amsfonts}       
\usepackage{nicefrac}       
\usepackage{microtype}      
\usepackage{xcolor}         
\usepackage{cite}
\usepackage{wrapfig}
\usepackage{graphicx}
\usepackage{float}

\title{P4AI: Approaching AI Ethics through Principlism}

\author{%
  Andre Fu\\
  University of Toronto\\
  \texttt{andre.fu@mail.utoronto.ca} \\
  \And
  Elisa Ding\\
  University of Toronto\\
  \texttt{elisa.ding@mail.utoronto.ca}\\
  \And
  Mahdi S. Hosseini\\
  University of New Brunswick\\
  \texttt{mahdi.hosseini@unb.ca}\\
  \And
  Konstantinos N. Plataniotis\\
  University of Toronto\\
  \texttt{kostas@ece.utoronto.ca}
}

\begin{document}

\maketitle

\begin{abstract}
 The field of computer vision is rapidly evolving, particularly in the context of new methods of neural architecture design. These models contribute to (1) the Climate Crisis - increased CO2 emissions and (2) the Privacy Crisis  - data leakage concerns. To address the often overlooked impact the Computer Vision (CV) community has on these crises, we outline a novel ethical framework, \textit{P4AI}: Principlism for AI, an augmented principlistic view of ethical dilemmas within AI. We then suggest using P4AI to make concrete recommendations to the community to mitigate the climate and privacy crises.
\end{abstract}

\section{Introduction}
The field of AI is experiencing rapid growth in terms of more accurate models and wider applicability. This is largely due to the design of new architectures and improvements on existing ones. Furthermore, state-of-the-art models are following the trend of higher resource usage in both energy and data. However, there are several risks involved in building larger models, namely the consequences of climate change and privacy violations. These crises are explored in detail with respect to CV, as well as key stakeholder groups in this context. In this work, we extend the work from \cite{fu2021reconsidering} and fully develop the concept of \textit{enforcement} via the ethical framework \textit{P4AI}: Principlism for AI. The concept of enforcement is proposed, which outlines actionable considerations to minimize the impacts of building and training neural architectures. Enforcement works as an extension of principlism, an ethical framework specialized for engineering applications. \cite{beever2016reflexive}. 

\section{Motivation}
AI and their neural architectures are becoming an ever-growing presence in our lives, as the global AI market is forecasted to reach \$299.64 Billion by 2026, at a growth rate of 35.6\% \cite{factsfactors_2021}. As society moves towards dependence on AI and ML technologies, the CV community has placed great significance on ethical principles around data biases, privacy, and algorithmic biases without much consideration towards the neural architectures that underpin these models. \cite{fu2021reconsidering} showed that CV models emit significant CO2 emissions throughout their lifespan, referenced in Appendix A for context. We highlight two crises whose causes can be directly addressed by neural architectures: (1) The Climate Crisis and (2) Privacy.

\textbf{Climate Crisis:}
The Intergovernmental Panel on Climate Change (IPCC) provides transparent and peer-reviewed reports on climate change for the scientific community to make informed decisions within their work. The 2019 IPCC Special Report proposed that limiting warming to 1.5$^{\circ}$C above pre-industrial levels is necessary to mitigate extreme climate-related risks. In 2016, we emitted 52 GtCO2 and by 2030 we will have 52-58 GtCO2, while we should be at 25-30 GtCO2 within the same time-frame. However, the recent IPCC Sixth Assessment Report shows that ``global warming of 1.5$^{\circ}$C and 2.0$^{\circ}$C will be exceeded in the 21st century unless deep reductions in CO2 and other greenhouse gas emissions occur in the coming decades'' \cite{ipcc2021}. Additionally, the report extrapolates to possible warming scenarios depending on the levels of greenhouse gases (GHG) emitted. Global surface temperatures during 2081-2100 will likely increase 1.0-1.8$^{\circ}$C under low GHG emissions, 2.1-3.5$^{\circ}$C under intermediate GHG emissions, and 3.3-5.7$^{\circ}$C under high GHG emissions \cite{ipcc2021}. These CO2 levels indicate that every person and community must play their part to curb the climate crisis, and it is urgent that we consider CO2 impact while creating and choosing neural architectures \cite{fu2021reconsidering}.




\textbf{Privacy Crisis:}
Our society is moving towards a highly interconnected system, where people and computers are being linked together to move information at extreme rates. As we become more interconnected, the question of privacy arises, in particular how can we defend privacy as a human right \cite{ohchrPrivacy} in the face of rising technology trends where data and privacy exploitation is common \cite{dataExploit}. Accordingly, inspecting the AI and ML models that underpin modern technology is essential in protecting privacy. As we inspect these models, it becomes apparent that model security and the neural architecture itself form the basis for privacy.

A framework that contextualizes model security and possible risks of breaches is the confidentiality, integrity, and availability (CIA) model \cite{chai_2021, samonas2014cia}. Compromised confidentiality may include leaking protected model details and private data (eg. medical records). Attacks on integrity result in biased outputs, for example by altering training data. Lastly, attempts to block access to model details are an example of compromised availability.

\subsection{Stakeholders} 
While both the climate crisis and privacy crisis are global dilemmas within modern society, we scope this to current researchers and  marginalized communities. In identifying these groups, we highlight the dynamic between the groups and the direct consequences they have on each other.
\begin{wrapfigure}{l}{0.35\textwidth}
    \centering
    \includegraphics[width=0.35\textwidth]{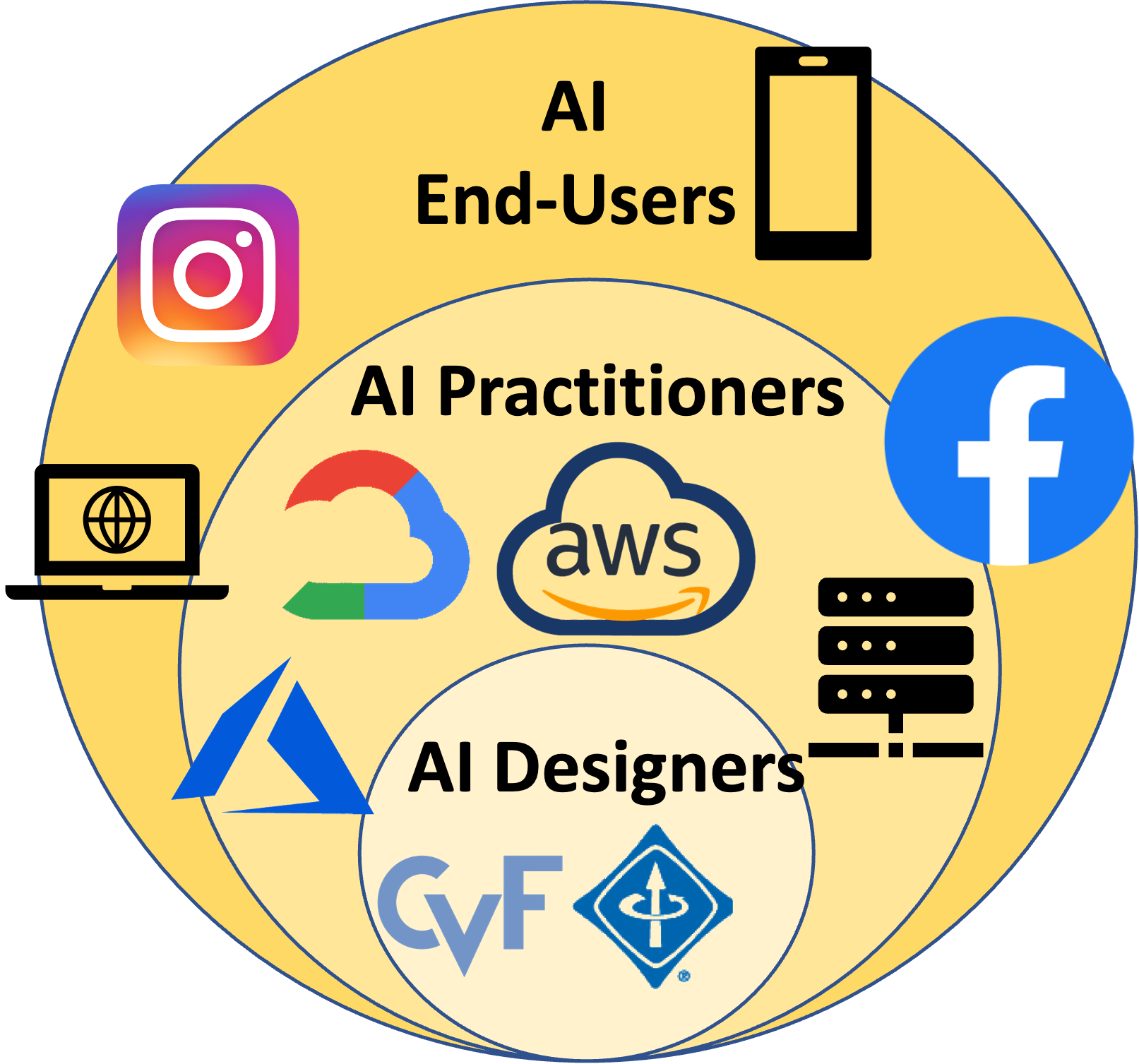}
\end{wrapfigure}

\textbf{Computer Vision Researchers:} Comprising the majority of the computer vision community, researchers play a direct role in combating the aforementioned crises. Our actions and novel architecture designs create not only long-term impacts on the rest of the community, but also downstream environmental implications, as seen in the figure on the left. As a researcher develops an architecture (AI designers), the impact propagates to other researchers, machine learning engineers (AI practitioners) and society-at-large \cite{fu2021reconsidering} (AI End-Users). Furthermore, the atmospheric CO2e emitted from the compute required for training and inference of deep learning models contributes to the climate stability and livelihoods of future peoples \cite{fu2021reconsidering}. Therefore, it is critical our community of computer vision researchers recognize our contributions to CO2 emissions.

Within the context of privacy protection, a plethora of concerns have risen around the computer vision and privacy risks. Recently, there have been discussions \cite{noorden2020facialrec} around facial occlusion reconstruction and the maleficent implications of using this technology to invade privacy, particularly in regions with oppressive regimes. Furthermore, computer vision models suffer from data-preservation gaps where models can be attacked where attackers can reconstruct \textit{private} training data. These threats are often not considered by CV researchers, and greater awareness should be shown for the downstream applications and implications of neural architectures.

\textbf{Marginalized Communities}: The actions taken by our community may have unintentional effects on marginalized communities, which we define as groups experiencing exclusion from mainstream social, economic, educational, and/or cultural life \cite{sevelius2020research}. As analyzed in \cite{fu2021reconsidering} in terms of CO2 emissions, they found a rarely discussed marginalized group: our future generations. They defined this group and highlighted a subgroup who would be most affected, those in equatorial and coastal regions. As CO2 emissions become an increasingly pressing issue, our future generations in equatorial regions will experience serious water-stress affecting between 75-250 million people in Africa. Furthermore, those in coastal regions will experience 1-8 feet of sea level increase and be at higher risk of losing their homes.

In terms of privacy, we particularly identify two marginalized populations:  those in oppressive regimes and peoples of colour. Deep-learning models are becoming increasingly popular to identify marginalized groups within oppressive regimes \cite{joplin, malik_2019} as they have the ability to efficiently target peoples. Furthermore, in the medical sector AI models can help diagnose those with niche diseases but via data-extraction techniques, patients can be personally identifiable through weak neural architecture design.

In identifying the key stakeholders, we demonstrated that neural architecture design considerations should extend beyond test-accuracy or performance, but also deeply into the ethical, societal and real-world. 

\section{Adding \textit{Enforcement} to Principlism}
Within the ethical AI community, we've adopted `Principlism' \cite{beauchamp2001principles, beever2016reflexive} as the primary framework to analyze ethical issues that arise \cite{floridi2018ai4people, jobin2019global, whittlestone2019role}. Principlism is an applied ethics approach encapsulating four pillars:
\begin{itemize}
    \item \textbf{Respect for Autonomy:} describes the requirement for an individual or group to be self-determining, specifically with no coercion 
    \item \textbf{Beneficence:} refers to the obligation to prevent harm and act in the benefit of others
    \item \textbf{Justice:} describes the requirement to ensure risks, costs and benefits are fairly distributed 
    \item \textbf{Non-maleficence:} entails avoiding causation of harm or avoiding actions that would cause intentional harm 
\end{itemize}
\begin{wrapfigure}{r}{0.25\textwidth}
    \centering
    \includegraphics[width=0.25\textwidth]{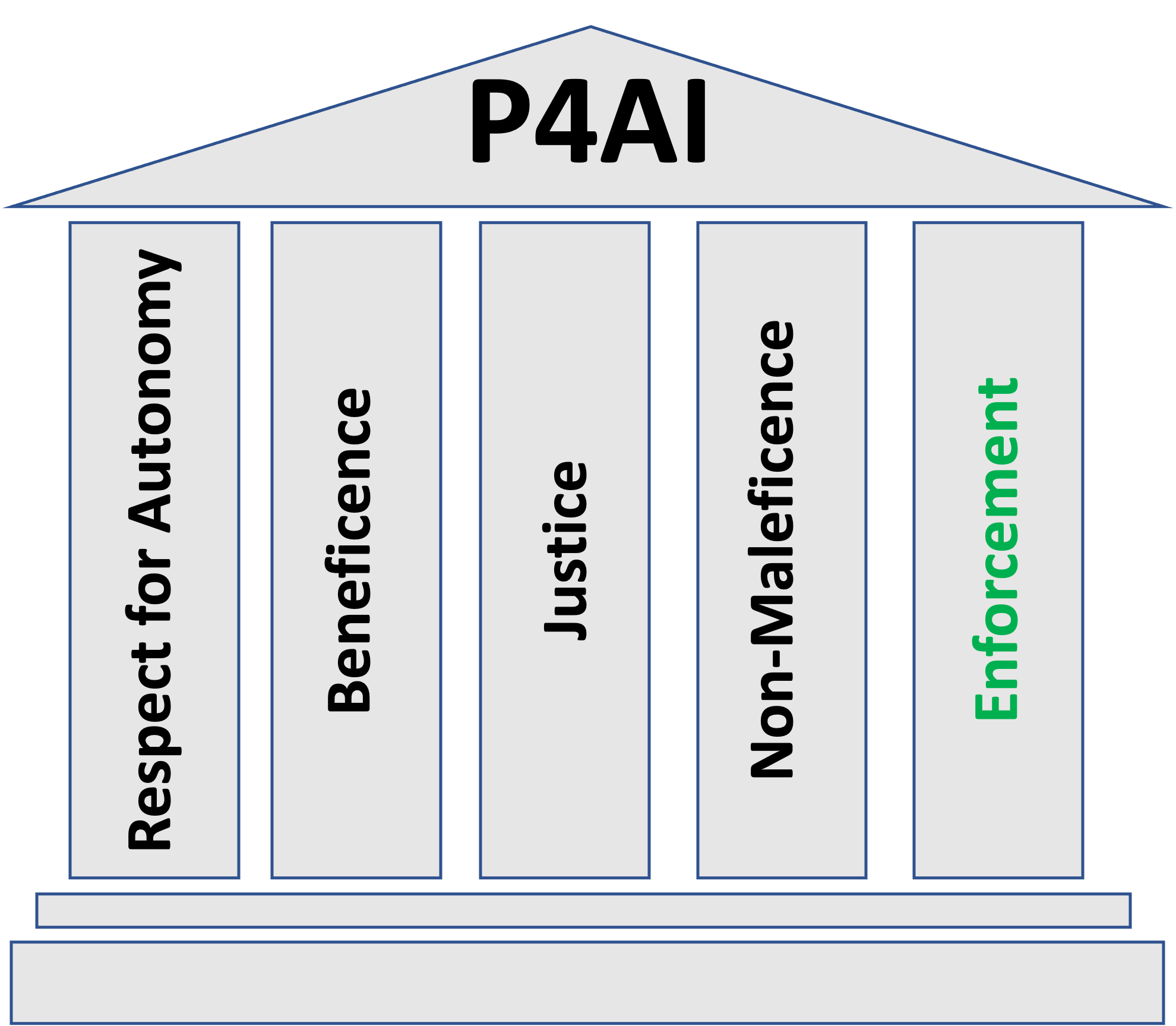}
\end{wrapfigure}

In this work, we propose \textit{P4AI}: Principlism for AI where we add a fifth pillar of ethical AI - \textit{enforcement}. Enforcement is the idea that our obligations to marginalized communities should be included within our research, furthermore these concepts should permeate our work. In realizing our impact, we can hold ourselves and our community accountable. 

\subsection{Applying Principlism For AI}
Through Principlism, we can analyze the impacts of designing neural architectures that do not consider ethical and down-stream implications. Here we apply the four principles through the lens of the climate and privacy crises. Two cases are discussed in Appendix B to describe possible ethical risks and enforcement as a method to reduce such risks.

\textbf{Climate Crisis:} As we emit this level of atmospheric CO2, we are overlooking future generation's capacity to be self-determining as we are forcing them to endure incredible amounts of water-stress, agricultural conflict and increased risk of extreme weather events. In contraposition to Beneficence, we are not acting in the benefit of future generations by actively emitting this level of CO2. As we currently enjoy a stable climate while marginalized groups and our future generations will experience the implications of our actions today, we're not fairly distributing costs, benefits an risks, violating Justice. Additionally, by causing harm to marginalized groups despite the signs of the climate crisis, we're actively causing harm thus violating Non-maleficence.\\

\textbf{Privacy Crisis:} Within the context of privacy, we are unintentionally violating Principlism by designing architectures that do not consider marginalized populations. We're inadvertently removing the ability for those in marginalized groups to be self-determining by targeting those in oppressive regimes. By creating neural architectures that are easily attacked, they can leak sensitive information thus causing harm, violating beneficence.  

\section{Design Considerations for Researchers}
\textbf{CO2 Reduction Methods: }
Applying enforcement in CO2 consideration involves reducing emissions in both the initial architecture search and lifetime use of models. FLOPs are the main established measure of a model's computational cost and can be incorporated as a target for minimization. Three main optimizations can be made to reduce subsequent environmental cost \cite{fu2021reconsidering}: 
\begin{itemize}
    \item \emph{NAS-RL}: building on hardware-aware NAS to include a CO2/FLOP objective
    \item \emph{Gradient-based}: including a CO2/FLOP constraint on the chosen-operations level of the bi-level
    \item \emph{Evolution}: adding a CO2/FLOP regularization parameter to the fitness function.
\end{itemize}
In addition to adopting these optimizations, researchers should strive to consistently report FLOPs values to build a culture of transparency and accountability. We can \emph{enforce} optimizing on FLOPS by refusing to use high FLOPs networks, thereby communally penalizing researchers publishing these types of models \cite{fu2021reconsidering}.

\textbf{Privacy Protection Methods: }
The extensive application of AI worldwide also introduces new threats of attacks on systems and their data. Measures should be taken to protect against attacks throughout all phases of design, training, and inference. The following are suggestions that guard against external manipulation \cite{papernot2016towards}:
\begin{itemize}
    \item \emph{Data poisoning detection: } Altered data can be detected through statistical approaches, as poisoned samples lie outside of the expected data distribution. For example, Rubenstein et al. developed a PCA detection model to identify and limit outliers of a training set \cite{biggio2011support}. Furthermore, introducing a regularization term to loss functions can reduce complexity \cite{biggio2011support}.
\end{itemize}

\textbf{Ethics Cards: }
Mitchell et al. \cite{mitchell} developed the framework of model cards for transparent reporting of machine learning models. These model cards provide documentation on the performance, datasets, and context of models. This includes information such as model details (developers, version/type, parameters, constraints), intended use, training/test data, and recommendations. We propose utilizing a similar technique to provide an accessible enforcement-based analysis to the community and public. These ethics cards should outline potential risks and promote transparent discussion regarding risks.

\textbf{Ethics Committee:} 
Currently, the process of publishing research in the computer vision community involves extensive peer review, often with a confidential and double blind process to ensure scientific integrity. However, there is no existing review process to ensure ethical integrity of AI research. We propose implementing an ethics committee to overview accepted papers and prepare a statement on the ethical considerations and risks of each paper.

\section{Conclusion}
We have defined the pressing concerns of the Climate and Privacy crises, as well as the role of AI and ML technologies. As researchers of these neural architectures, we hold strong responsibility for minimizing their CO2 emissions and associated privacy risks. This is further driven by consideration of marginalized communities who cannot defend themselves against the crises. Enforcement was introduced as an extension of Principlism, where actionable steps must be upheld by peers to achieve universally high ethical standards. Enforcement methods promote CO2 reduction and data protection directly during the development of neural architectures in addition to transparency of these architectures. Example cases were also presented to show how climate and privacy effects could be mitigated in prior work. We hope that enforcement can be implemented within the community to aid in developing positively impactful neural architectures. 

{\small
\bibliographystyle{ieeetr}
\bibliography{egbib}

\begin{thebibliography}{10}
\bibitem{joplin}
T.~Joplin, ``Artificial intelligence isn't an agent of oppression but in china
  it's already the tool.''
\end{thebibliography}
}

\newpage

\appendix

\section{Lifetime CO2 Impact of Major Models}
\begin{figure}[H]
    \centering
    \includegraphics[width=0.5\textwidth]{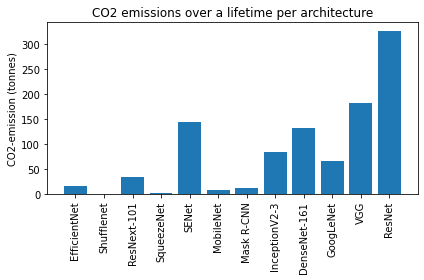}
    \caption{CO2 emissions of hand-crafted models over their lifetimes so far. The common architectures are ordered from left-to-right as lowest-to-highest number of citations [1]}
\end{figure}

\begin{table}[H]
\centering
\begin{tabular}{c|c|c|c}
                   & \textbf{CO2 (tons)} & \textbf{\begin{tabular}[c]{@{}c@{}}Cars \\ Driven\end{tabular}} & \textbf{\begin{tabular}[c]{@{}c@{}}Homes \\ Powered\end{tabular}} \\ \hline
\textit{ResNet} \cite{he2016deep}   & 326.6               & 70.6                                                            & 55.3                                                              \\ \hline
\textit{VGG} \cite{simonyan2014very}& 181.7               & 39.8                                                            & 30.8                                                              \\ \hline
\textit{GoogLeNet} \cite{szegedy2015going} & 65.1                & 14.1                                                            & 11                                                               
\end{tabular}
\vspace{0.2cm}
\caption{Contextualized CO2 costs from the 3 most popular hand-crafted models with cars driven \& homes powered over 1 year. \cite{fu2021reconsidering}}

\end{table}

\section{Example Cases}

\textbf{Facial Reconstruction: }
New techniques have been developed using encoder-decoder architectures to fill masked regions, particularly for the application of facial reconstruction \cite{li2017generative}. Models are able to generate realistic facial features despite varying types of occlusions in the input images. The end uses of these architectures may forensics, identification of missing persons, and surgical reconstruction planning. However, there are a multitude of risks regarding the privacy of individual faces. This violates the confidentiality aspect of the CIA security model. In cases where it is impossible to anonymize data (eg. blurring faces), express consideration must be given to consent and ethical sourcing of data. In this scenario, researchers should also strive to produce ethics cards addressing these risks. By building a culture of enforcement within the community, our progress will be directed towards improving society at large as opposed to producing case-specific results.

\textbf{AlphaZero/AlphaGoZero: }
Over the past decade, innovations in algorithm design, data, and hardware have accelerated the amount of compute power used in AI. AlphaZero and AlphaGoZero are two breakthrough models in AI \cite{silver2017mastering}, but both require approximately 1e3 petaflop/s days to train (1 pfs-day = $10^{15}$ operations per second for one day) \cite{amodei_2021}. This is a 300,000x increase in compute compared to older models such as AlexNet \cite{amodei_2021}. Furthermore, from 2012 to 2018 alone, the doubling time of pfs-day usage is approximately 3.4 months indicating exponential growth \cite{amodei_2021}. Despite the accomplishments and historical impact of AlphaGo in society, compute cost should be considered during the design process and minimized where possible. Optimizations such as AI-specific chips and parallelizing operations help increase energy efficiency, but they also have limitations and do not solve the problem entirely. The contributions of new work should be weighed against its overall cost of production. Enforcement can be applied with respect to the climate crisis by researchers collectively making efforts to reduce emissions of their work, which will ensure more sustainable development within the field.

\end{document}